\documentclass[12pt, a4paper]{article}

 \usepackage[bottom]{footmisc}% places footnotes at page bottom

\usepackage{graphicx}
\usepackage{amsthm}   % See h.c. printout for how to use this
\usepackage{amsmath} % assumes amsmath package installed
\usepackage{amssymb}  % assumes amssymb package installed
\usepackage{mathrsfs} % for Lagrangian symbol
\usepackage{stmaryrd} 
\usepackage{txfonts} % for varprod (Cartesian product)

\usepackage{hyperref}  % For embedded hypertext
\usepackage[capitalize]{cleveref}

\usepackage{enumerate}
\usepackage{appendix} 

\usepackage[all]{xy}

\newcommand{\mathsym}[1]{{}}
\newcommand{\unicode}[1]{{}}

\newcommand{\E}{{\mathbb{E}}}
\newcommand{\R}{{\mathbb{R}}}
\newcommand{\B}{\{0,1\}}

\newcommand{\Z}{{\mathbb{Z}}}

\newcommand{\A}{{\mathcal{A}}}
\newcommand{\X}{{\Phi}}
\newcommand{\XX}{ {\hat{\Phi}} }
\newcommand{\I}{{\mathcal{I}}}
\newcommand{\II}{\hat{\mathcal{I}}}

\usepackage{color}  % for colors
\RequirePackage[dvipsnames,usenames]{xcolor}

\usepackage[obeyFinal]{todonotes}
\usepackage{ifthen,xspace}

\newcommand{\eq}[1]{\begin{align}#1\end{align}}

\newcommand{\be}{\begin{equation}}
\newcommand{\bel}[1]{\begin{equation}\label{#1}}
\newcommand{\qe}{\end{equation}}
\newcommand{\ee}{\end{equation}}
\newcommand{\eeq}{\end{equation}}
\newcommand{\ba}{\begin{eqnarray}}
\newcommand{\ea}{\end{eqnarray}}

\def\bal#1\eal{\begin{align}#1\end{align}}
\def\bann#1\eann{\begin{align*}#1\end{align*}}

\begin{document}

\title{The Implications of the No-Free-Lunch Theorems for Meta-induction}

\author{David H. Wolpert \\
 Santa Fe Institute, 1399 Hyde Park Road, Santa Fe, NM, 87501\\
 \texttt{http://davidwolpert.weebly.com}
 }

\date{}

\maketitle

%\end{document}

\begin{abstract}
The important recent book by G. Schurz ~\cite{schurz2019hume}
appreciates that the no-free-lunch theorems (NFL) have major implications for the problem of (meta) induction. Here I review the NFL theorems, emphasizing that they do not only concern the case where there is a uniform prior --- they prove that there are ``as many priors'' (loosely speaking) for which any induction algorithm $A$ out-generalizes some induction algorithm $B$ as vice-versa. 
Importantly though, in addition to the NFL theorems, there are many \textit{free lunch} theorems. In particular, the NFL theorems can only
be used to compare the expected performance of an induction algorithm $A$, considered in isolation, with the expected
performance of an induction algorithm $B$, considered in isolation. There is a rich set of free lunches which instead concern the
statistical \textit{correlations} among the generalization errors of
induction algorithms. As I describe, the meta-induction algorithms that Schurz
advocates as a ``solution to Hume's problem'' are simply examples of such a free lunch based on
correlations among the generalization errors of induction algorithms. I end by
pointing out that the prior that Schurz advocates, which is uniform over bit frequencies rather than bit patterns,
is contradicted by thousands of experiments in statistical physics and by the great success of the maximum entropy
procedure in inductive inference.
%
%To illustrate the implications, consider choosing whether to use some fixed algorithm A or some fixed algorithm B to generalize from any given training set d, based on which algorithm has better cross-validation error on d. That use of cross-validation is itself a way to generalize from d, i.e., it is a learning algorithm. Call it C. As an alternative, consider algorithm D, which chooses whether to generalize from d using algorithm A or B based on which has worse cross-validation error on d. Arguably, C is a caricature of meta-induction, while D is a caricature of “anti-meta-induction”. NFL tells us that there are as many prior for which anti-meta-induction outperforms meta-induction as vice-versa.
\end{abstract}

$ $

$ $

\noindent ``There is only limited value

\noindent in knowledge derived from experience.

\noindent The knowledge imposes a pattern, and falsifies,

\noindent for the pattern is new in every moment.''

--- T.S. Eliot

\section{Introduction}   
\label{sec:introduction}

Inductive inference plays a central role in an extraordinarily wide variety of fields, ranging from traditional
statistics to Monte Carlo estimation~\cite{rubinstein2016simulation,tracey2013using} to community detection and link detection in networks~\cite{peel2017ground,ghasemian2020stacking,guimera2020one}. It is also central
to essentially all flavors of machine learning~\cite{bishop2006pattern,kroese2019data}, ranging from supervised
learning (in which one is provided samples of an input-output map and wishes 
to infer the full map) to unsupervised learning (in which one is provided samples of 
a probability distribution and wishes 
to infer the full distribution) to active learning / experiment design (in which one is provided samples of an input-output map and wishes 
to determine what input to sample next, ultimately in order to infer the full map from all the samples) to reinforcement
learning (in which one is provided samples of an action-reward map and wishes 
to infer a sequences of actions that will maximize the expected discounted sum of rewards).
% to transfer learning 
%(in which the algorithm is provided samples of a probability distribution
%over a space and wishes 
%to infer the full distribution) to online learning, etc. 

The fact that inductive inference is central to all these fields
means that we can sometimes use it as a dictionary, to ``translate'' techniques developed in one field over to another field, providing
novel, powerful tools in that second field. As an example I am personally familiar with, the technique of `stacking'
is a powerful meta-inductive algorithm 
introduced in machine learning and statistics~\cite{brei96,clar03,smyth1999linearly} where it
is still being actively investigated and extended~\cite{yao2018using}. It can be 
translated from those fields into the field of Monte Carlo estimation. In that
new setting, stacking becomes a technique for post-processing a set of Monte Carlo
samples of a distribution to improve the accuracy of the associated estimate of an expectation value. Empirically, this use of stacking
seems to improve the accuracy of the estimator
\textit{no matter what precise Monte Carlo sampling algorithm is used} to generate the samples (simple sampling, importance sampling,
quasi-Monte Carlo, etc.). As another example, stacking has been translated to the domain of link prediction
in network science. In that domain it recently outperformed $203$ alternative algorithms by optimally combining
them, without making any Bayesian assumptions concerning the relative merits of those
algorithms~\cite{ghasemian2020stacking,guimera2020one}.

%\dhwc{Describe stacking's use for community detection. Also cite Gelman / Yuling recent work.}

Another illustration of how inductive inference techniques are shared among multiple  
fields is provided by one of the pillars of the scientific method. Science relies deeply on the assumption
that  one can inductively infer future 
predictive accuracy from current \textit{out-of-sample} predictive accuracy, even if as Hume emphasized, 
current in-sample predictive accuracy cannot be used that way~\cite{hume2003treatise}.  To be a bit
more precise, suppose we have
a set of scientific theories, $T = \{T_i\}$, each of which can make predictions about the outcomes
$y \in Y$ of any one of a set of experiments $x \in X$. Let $d$ be the set of all experiment-outcome pairs
that were observed by the people who made those theories. Let $x_{-d}, x'_{-d}$
be two experiments that are not in $d$. A central assumption of the scientific
method is that whichever of the theories in $T$ is most accurate when predicting
the outcome of the experiment $x_{-d}$ would likely be more accurate than the other theories when predicting
the outcome of the experiment $x'_{-d}$. In other words, if we choose among theories based
on a set of new experiments $S$ which were not used to create those theories in the first place, then we are likely to find
the theory which would also be most accurate in all future experiments we might conduct after the ones in $S$. 

If we translate this assumption underlying the scientific method into the fields of
machine learning and statistics, we end up with the 
technique of cross-validation, which is a core component of those fields.\footnote{
To see this relationship, note that cross-validation chooses among a set
of learning algorithms (rather than theories), and does so according to which of those
performs best at out-of-sample prediction (evaluating that performance
by forming ``folds'' of the single provided data set). }
%\dhwc{complete this; out-of
%sample prediction as a way to choose among theorists; very successful; can be extended 
%to give machine learning technique of cross-validation, which has also proven stunningly successful ...n } 
In turn, the technique of cross-validation
can be translated from the field of machine learning to the field of ``black box'' optimization. In that new field, it provides a way
to dynamically set the parameters of an optimization algorithm (e.g., to dynamically set the temperature in simulated annealing).
Just as out-of-sample induction has proven so successful in science in general, and was then
so successful when translated into machine learning in the form of cross-validation, 
it has also been found empirically that when translated into the domain of black box optimization it
results in faster convergence of the optimizer to the global optimum 
solution~\cite{wolpert2013using,adam2019no}. 

Despite sharing inductive inference as a central component, there are many important distinguishers
among these fields; they are not simply different expressions of the same underlying phenomenon. 
In particular, the no-free-lunch (NFL) theorems apply to to optimization~\cite{woma97} and to certain aspects of
both supervised learning~\cite{wolp96a} and community detection~\cite{peel2017ground}. However, they do not apply to 
other aspects of supervised learning~\cite{wolp96b} and do not apply to co-evolutionary 
optimization~\cite{woma05}; there are ``free lunches'' in those domains.
%They also do not apply to many types of ``online learning ''.

Another example of a free lunch was introduced in 1996 by Parrondo and colleagues~\cite{parrondo1996criticism,harmer1999losing}.
Suppose one has a pair of games that an agent can play, where the agent has higher probability of losing than winning
in both games. Parrondo devised a strategy for the agent to the play those games in an alternating sequence
which results in \textit{higher} probability of winning, despite the fact that in each game, separately, in each round of
play, the agent faces a higher probability of losing. This result was called ``Parrondo's paradox'', since it was so surprising 
when it was discovered. As mentioned above, Parrondo's result can be viewed as a ``free lunch'', related to the meta-induction algorithms
considered by Schurz. (As an historical aside, it's interesting to note that Parrondo went on to make some of the seminal 
contributions to stochastic thermodynamics and non-equilibrium statistical physics~\cite{parrondo2015thermodynamics}).

To properly understand the claims made in the recent book by Schurz~\cite{schurz2019hume}, it's worth taking
a moment to walk through a simplified version of the scenario that Parrondo analyzed.
% Fix some initial
%iteration $t_0$, and suppose that every iteration $t$ thereafter, you move a large fraction of your total stake into the stock that has done best
%on aggregate during the interval $[t_0, t]$. If you chooses that fraction in an intelligent manner, then it is
%unlikely that you will incur large regret (i.e., that your total winnings will be substantially worse than the best they could have been,
%if you had magically known how the stock market will behave). This is true independent of the stochastic process
%governing the dynamics of stock prices.
%
%To understand intuitively why this kind of stock-picking strategy might limit expected losses, as a simplified version of it, suppose
Suppose
there are $K + 1$ infinite sequences of bits, $\{v_k(i) \in \{0, 1\} : k = 1, \ldots, K+1, i \in \Z^+\}$.
The first $K$ of those sequences are the successive payoffs that a player would have received if they had
picked that sequence. So
for any counting number $n$, the accumulated payoff the player would have by iteration $n$ of
the sequences if they had always picked sequence $k$ is 
$\pi_k(n) := \sum_{i=1}^n v_k(i) $. Define the ``best'' sequence on a given iteration $n$ as $k^+(n) := 
\arg \max_{k} \pi_k(n)$, and the ``worst'' one as $k^-(n) := \arg \min_{k} \pi_{k}(n)$. 

Next,  suppose that on any iteration $n$, the associated bit $v_{K+1}(n)$ of the $K+1$'th sequence is the value
given by choosing the sequence whose accumulated payoff over the \textit{previous} $n-1$ 
iterations was highest, and evaluating its payoff bit for iteration $n$.
Formally, $v_{K+1}(n) = v_{k^+(n-1)}(n)$. 
%Under this ``all-or-nothing'' rule for $v_{k+1}(n)$,
%at the end of each iteration $n$
%one moves all of one's current holdings into the single stock that cumulatively performed best in the preceding 
%$n-1$ iterations, independent of whether you ever owned that stock previously.
%%It models the simplified version of) Parrondo's strategy.
This rule for how to choose among the sequences on each successive
iteration  is a simplified version of Parrondo's strategy. 

To see why it can be good to follow this strategy, consider the limiting case where
$K = 2$. Suppose that on some iteration $n$ the difference $\pi_{k^+(n)}(n) - \pi_{k^-(n)}(n) = n$. Then it must that
in \textit{every} iteration $i \le n$, $\pi_{k^+(i)}(i) - \pi_{k^-(i)}(i) = i$ (since the maximal difference in any
single iteration is $1$). So either both $v_1(i) = 1$ and $v_2(i) = 0$ for all $i \in \{1, \ldots, n\}$, or vice-versa.
%Therefore, it must in fact be the case that for all $m \le n$, $\pi_{k^+(m)}(m) - \pi_{k^-(m)}(m) = m$.
As a result, $v_{K+1}(i) = i$ for all those iterations. In other words, following the strategy $v_{K+1}$ results in
zero ``regret'', in the formal sense of the word, in all iterations before $n$ --- there will be zero gap between one's actual
accumulated payoffs and  the best possible value of the accumulated payoff,
which one could have achieved if one knew the
entire sequence of all payoffs for all time before picking among those sequences.  Formally,
$\pi_{K+1}(n) = \pi_{k^+(n)}(n)$.

If instead $\pi_{k^+(n)}(n) - \pi_{k^-(n)}(n)$ is some large value, but not quite $n$, then we can still provide guarantees
that in \textit{most} of the steps $m < n$, following the strategy $v_{k+1}(m)$ would have resulted in \textit{very little} regret.
For example, if $\pi_{k^+(n)}(n) - \pi_{k^-(n)}(n) = n - 2$, then in exactly one of the iterations $i \le n$, the payoff
$v_{k^+(n)}(i) = 0$ and $v_{k^-(n)}(i) = 1$, while in all other iterations, the opposite is true. So for all iterations $j < i$,
there will be zero regret for using sequence $K+1=3$, while for all iterations $j \ge i$, there will be regret $2$ for
using that sequence. So if $\pi_{k^+(n)}(n) - \pi_{k^-(n)}(n) = n - 2$, then
we are guaranteed that the regret never exceeds $2$ in any iteration $i \le n$, if we use strategy $K+1$, and that in half of all
sequences, there would have been zero regret for using that strategy for at least half of the $n$ iterations.

As $\pi_{k^+(n)}(n) - \pi_{k^-(n)}(n)$ shrinks, those guarantees on the fraction of iterations with little regret for using strategy $K+1$
get weaker --- but in addition, the maximal regret for using that strategy
shrinks. So in general, following the strategy $v_{K+1}(i)$ for all iterations will not frequently result in a large amount of regret.
In contrast, choosing some specific strategy $M \in \{1, 2\}$ and using that strategy for
all iterations could result in quite bad regret by the iteration $n$. Conversely, if it \textit{were} the case that
using some such specific strategy $M \le K$ resulted in little regret, then it would also be the case that using
strategy $K+1$ would result in little regret.
In this sense, strategy $v_{K+1}(i)$ is superior to the other two strategies, 
with payoffs $v_1(i)$ and $v_2(i)$ respectively, \textit{no matter what the sequences $\{v_k(.)\}$ are}. 

As $K$ grows, this guarantee gets weaker --- but
it always holds. Moreover, there are strategies $v_{K+1}$ that at each step use the preceding sequences
$\{v_k(.)\}$ in a more nuanced way than the all-or-nothing rule $v_{K+1}$ described above. (In particular,
the strategy underlying the Parrondo paradox is more nuanced than the all-or-nothing rule.) Importantly,
all of these results hold even though no underlying probability distributions have been specified. 
%\dhwc{If there were probability distributions, and the process generating payoffs at each iteration were a martingale,
%would that guarantee that there is no possible strategy for setting sequence $K+1$ that can result in expected
%accumulated payoff by any iteration $n$ that is better than the expected payoff of uniformly randomly
%choosing among the sequences?}

In some senses, one might argue that this Parrondo-like strategy for picking among sequences
had a precursor in the informal investigations of Reichenbach~\cite{reichenbach1938experience}. 
Related work was also done in 1997~\cite{cesa1997use}. (See also~\cite{cesa1996worst}.)
That particular analysis was later elaborated and extended in 2006~\cite{cesa2006prediction}. The general topic of
designing and analyzing algorithms for these kinds of scenarios is now known 
as ``online learning under expert advice'' (OLEA). 
%Note that as interesting as OLEA is,
%the associated guarantees are quite weak. In particular, very few stock traders (none?)
%use any of the associated algorithms, despite those guarantees.

Schurz claims in~\cite{schurz2019hume},
that properly elaborated, these OLEA results ``justify (meta-)induction'' and ``solve Hume's problem''.
In particular, such claims are made on behalf of various \textbf{attractivity-weighted} (AW) algorithms.
Of course, since Schurz only considers the
(rather limited) version of induction addressed by OLEA, these results cannot be said to ``justify induction''
in the full sense of inductive inference discussed at the beginning of this chapter. 
Moreover, since as mentioned above it was already known that there are {free}-lunch 
theorems~\cite{woma05,wolp96b}, some forms of (meta-)induction already had been ``justified''.
%and (at most) Schurz is elaborated this justification. (Note that it was even already known that, in certain senses, there
%are free lunches in supervised learning --- see~\cite{apriori}.)

It is also important to emphasize that in actual scientific practice, theories
are not continually revised with each new experimental datum  --- in the language of the example above, $v_{K+1}(n)$
is modified to enforce the constraint that it only changes what sequence it chooses quite infrequently, rather than at every iteration,
as in its original version described above. Similarly, in actual scientific practice, 
at any given time scientists are only considering at most a few theories --- in the language of the example
above, $K$ is quite small. (After all, it would simply be too expensive, in many
different aspects, to have many different scientific theories all continually being updated.) For these and other reasons,
the OLEA guarantees are quite weak when applied to actual scientific practice. So they provide
little justification for the kind of induction scientists use.

This still leaves open the possibility that the kinds of guarantees given by OLEA
could be significant in an idealized version of scientific practice.
In this chapter I further analyze Schurz's claims that this is true, in light of the NFL theorems.

%On to 
%
%Note that supervised learning is not a ``prediction game'', in the language of \cite{schurz2019hume}. A prediction game
%would involve an infinite stream of queries, and therefore (for finite $X, Y$) result in repeated queries, either all OTS
%or some on-training set.

\section{The no free lunch theorems}

The NFL theorems for supervised learning are the ones most relevant for discussions of ``induction''
in the sense meant by Schurz. 
Let $X$ be a finite \textbf{input} space, $Y$ a finite \textbf{output} space. Suppose we have 
a \textbf{target distribution} $f(y_f \in Y \mid x \in X)$, along with 
a \textbf{training set}  $d = (d^m_X, d^m_Y)$  of $m$ pairs $\{(d^m_X(i) \in X, d^m_Y(i) \in Y)\}$, that
is stochastically generated according to a distribution $P(d \mid f)$ (a conditional distribution conventionally called a
\textbf{likelihood}).
%the rule
%\begin{eqnarray}
%P(d \mid f, m) \propto \rho(d^m_X) \prod_{i=1}^m  f(d^m_Y(i) \mid d^m_X(i))
%\end{eqnarray}
%where $\rho(.)$ is a (potentially non-IID) \textbf{sampling probability} of any set of $m$ elements from $X$. 
Assume that based on $d$ we have a \textbf{hypothesis distribution} $h(y_h \in Y \mid x \in X)$.
The creation of $h$ from $d$ is completely arbitrary. It is specified \emph{in toto} by the distribution $P(h \mid d)$, and
is conventionally called the \textbf{learning algorithm}.
In addition, let $L(y_h, y_f)$ be a \textbf{loss function} taking $Y \times Y \rightarrow \R$.
Next, fix some distribution $P(q \,|\, d_X)$. Finally, given these distributions, define the associated \textbf{cost function} by
%\begin{eqnarray}
\eq{
C(f, h, d) \propto \sum_{y_f \in Y, y_h \in Y} \sum_{q \in X} P(q | d_X) L(y_f, y_h) f(y_f \mid q) h(y_h \mid q)
\label{eq:1}
}
%\end{eqnarray}

The cost function quantifies how well the algorithm does, averaged over all query points $q$, when the target is $f$,
the hypothesis generated by the algorithm is $h$, and the training set is $d$.
Note that the term $P(q | d_X)$ in the definition of the cost function governs how a query point $q$ is generated for testing the performance
of the algorithm, given the set of points the algorithm has already seen. So for example, under IID sampling to generate both
the query point and the training set, $P(q | d_X)$ is independent of $d_X$. In contrast, if 
we are concerned with the ability of the algorithm to \textit{generalize} from the training set, then we might require
that $P(q | d_X) = 0$ if $q \in d_X$, since if the query point were the same as an element of the training set, then the cost function
would quantify the memorization performance of the algorithm, not the generalization performance.
%where $P_{d_X}(q)$ is some function of $q$ that may (or may not) vary with $d_X$.

From now on, for simplicity, I will assume that any $f$ is a single-valued function (i.e., $f(y_f \,|\, x)$ is a delta
function for each $x$) and similarly for any $h$. I will also assume that the training-set generation process
is ``vertical'', in the sense that $P(d_Y \,|\, d_X, f)$ is independent of the values of $f(x)$ for $x \not \in d_X$.

%All aspects of any supervised learning scenario --- including the prior, the learning algorithm, the
%data likelihood function, etc. --- are given by the joint distribution $P(f, h, d, c)$ (where $c$ is values of the cost function)
%and its marginals.
As an example of this framework, in Bayesian statistics analyses of ``model mis-specification'', 
one might investigate the posterior expected cost,
\eq{
\E(C \,|\, d) &= \sum_{f, h} P(h \,|\, f, d) P(f \,|\, d) C(f, h, d) \\
	&= \sum_{f, h} P(h \,|\, d) P(f \,|\, d) C(f, h, d)
\label{eq:2}
}
(the second line following from the fact that the hypothesis generated by the learning algorithm is conditionally independent of the target,
given the training set). In contrast, in sampling theory statistics and computational learning theory, one
is typically interested in
\eq{
\E(C \,|\, f, m) &= \sum_{h, d} P(h \,|\, d) P(d \,|\, f, m) C(f, h, d)
\label{eq:3}
}
where $m$ is the size of the training set.

This set of definitions is known as the ``extended Bayesian framework (EBF)''~\cite{wolp95b}.
The EBF is needed to properly understand the relationship between Bayesian and non-Bayesian statistics. It also
allows us to go beyond those two bodies of work. For example, we can use the EBF to
analyze the conditional distribution $\E(C \,|\, m)$. This allows us to derive a Bayesian
correction to the conventional bias-plus-variance decomposition that arises in sampling theory statistics~\cite{wolp97,kohavi1996bias}.
(The interested reader is directed to~\cite{wolp95b,adam2019no} for 
further-ranging discourse on how to 
integrate Bayesian and non-Bayesian statistics into an overarching probabilistic model of induction.)

Often when computational learning theory researchers refer to the ``generalization error'' of a 
supervised learning algorithm, they have in mind a \textbf{data-blind} cost function, meaning they choose
$P(q \,|\, d_X)$ in \cref{eq:1} to be independent of $d_X$. (For example, often one
assumes that $d_X$ was formed by IID sampling a distribution $\pi(x)$, and that $P(q \,|\, d_X) = \pi(q)$.) However, this choice
of $P(q \,|\, d_X)$ conflates two very different aspects of induction: being able to recall elements of the training
set (i.e., cases where $q \in d_X$), versus truly ``generalizing'' from the training set, to previously unseen instances (i.e., 
cases where $q \not \in d_X$). 

To help disentangle these two aspects of induction, one needs to use a distribution $P(q \,|\, d_X)$ 
that has zero measure on $d_X$, to focus on the generalization. Any $C(f, h, d)$ with
this choice is known as an \textbf{off-training set (OTS)} 
cost function, and generically written as $C_{OTS}(f, h, d)$. 
The key feature of an OTS cost function is that it only depends on the partial functions
$\{f(x) : x \not \in d_X\}$ and $\{h(x) : x \not \in d_X\}$. (See~\cite{wolp96a})A standard example is any function of the type
\eq{
C_{OTS}(f, h, d) &= \dfrac{\sum_{q \not \in d_X} \pi(q) L(f(q), h(q))}{\sum_{q \not \in d_X} \pi(q)}
\label{eq:OTS_example}
}
%for some loss function $L(.)$, where $\pi(.)$ is called the ``sampling distribution''.
In the words of Schurz~\cite{schurz2019hume}, ``the ultimate
goal and evaluation criterion of inductive inferences is success in predictions.'' Taken literally, this
would imply that we should \textit{only} be interested in OTS error.
%, and will be written as $P_{OTS}(q \,|\, d_X)$. 
My personal view is that OTS cost is neither a ``right'' or ``wrong'' way to measure performance --- rather it is
an analytic tool for distinguishing two very different properties of any learning
algorithm.

The no-free lunch theorems are also an analytical tool, designed to disentangle what aspects of 
a given learning algorithm can provide \textit{a priori} guarantees concerning its expected OTS cost.
It does by proving that if any given learning algorithm has particularly good OTS cost for one set of target functions,
it must have correspondingly \emph{poor} OTS cost on all other target functions. Formally, one of the NFL theorems says that
for a broad range of choices of loss function (formally, for any ``homogeneous loss function''),
for any likelihood function, $\sum_f P(C_{OTS} \,|\, f, m )$
is independent of the learning algorithm. Similarly, another NFL theorem says that if $P(f)$ is uniform, then
$P(C_{OTS} \,|\, d)$ is independent of the learning algorithm. These two NFL theorems imply
in particular that whether one uses the the type of expected value of interest in Bayesian statistics (\cref{eq:2}) or the one
of interest in non-Bayesian statistics (\cref{eq:3}), there are no \textit{a priori}, assumption-free benefits to using one
learning algorithm rather than another one.

%The NFL theorems tell us that
%if any learning algorithm performs particularly well on one set of target functions,
%it must perform correspondingly \emph{poorly} on all other target functions.
A secondary implication of the NFL theorems is that {if} 
%it so happens that 
you \textit{do} make an assumption, but it's that 
$P(f)$ is uniform, then the {average over $f$'s used in the NFL theorem
is the same as $P(f)$. In this case,}
you must conclude that all learning algorithms perform equally well{ for your assumed $P(f)$}.
This second implication is only as legitimate as is the assumption of uniform $P(f)$ it is based on, of course.
%Once other $P(f)$'s are allowed, the implication need not hold.
%Of course, this is true for all other arguments for using one search algorithm rather than another,
%which must (either explicitly or implicitly) make some assumption for $P(f)$.

However, it must be emphasized that simply allowing $P(f)$ to {be non-}uniform, \emph{by itself}, does not 
invalidate the NFL theorems. {Arguments that only say that $P(f)$ is non-uniform in the real world, without
advocating one particular non-uniformity,
do not establish anything whatsoever about what learning algorithm to use in 
the real world. In fact,  allowing $P(f)$'s to vary provides us with a new NFL theorem. In this
new theorem, rather than compare the performance of two learning algorithms by uniformly averaging over all $f$'s, we
compare them by uniformly averaging over all $P(f)$'s. The result is what one might expect: If any given algorithm
$A$ performs better than algorithm $B$ over a given set of $P(f)$'s, then it must perform corresponding worse
on all other $P(f)$'s. 

This is the main message of the NFL theorems, not the fact that inference is impossible under the uniform prior $P(f)$.
In fact, whether the uniform prior is ``induction-hostile'' is in fact irrelevant. The NFL theorems
do \textit{not} assume that the universe is governed by a uniform prior in some objective sense.
Nor do they suppose that the uniform prior somehow best captures our subjective ignorance about the universe ---
the NFL theorems do not motivate the uniform prior by invoking some variant of the common ``maximal ignorance'' reasoning underlying various
priors found in the Bayesian statistics literature.
%
%This does not mean that all algorithms 
%In particular, if algorithm $A$ outperforms algorithm $B$ for the uniform prior, then $B$ must outperform
%$A$ (on average) over all other priors. In this, the uniform prior is just like any other prior.
%There is no sense in which the uniform prior is in some sense special because it is so ``induction-hostile''; if it
%is induction-hostile to algorithm $B$, then it is ``induction-friendly'' to algorithm $A$, and there is some other prior
%that is induction-hostile to algorithm $A$.
%
%In fact, the NFL theorems hold when the state-uniform prior is maximally \textit{un}privileged, since they hold under a uniform distribution
%over all priors.
%More importantly, to 

To re-emphasize, the NFL theorems are a mathematical \textit{tool}, for analyzing \textit{a priori}
relationships between learning algorithms. It is a category error to interpret them as based on any ``epistemic assumptions''.
Indeed, what they force us to do is try to construct very weak assumptions that are not only reasonable,
but also can be exploited to design learning algorithms that perform better than random guessing.  
(See~\cite{wolpert1990relationship} for earlier work in this vein.)
Summarizing, Schurz is simply wrong when he states, ``Wolpert seems to assume that the state-uniform prior 
distribution is epistemically privileged.'' (\cite{schurz2019hume}, p. 240) --- the NFL theorems make
no assumption whatsoever concerning the epistemic nature of the uniform prior. 

I end this section by emphasizing that NFL is completely consistent with many
%this same uniform $P(f)$ can also be used to establish many
\textit{free lunches}. Crucially, the NFL theorems equate the expected OTS performance
of any two learning algorithms \textit{only when they are considered independently}, in isolation from one another. However, in
general the OTS performance of any two algorithms can be correlated as one varies over $f$'s. 

As an example, depending on the likelihood, loss functions, and other details, it may
be that for all $f$, the expected OTS error of algorithm $A$, $\E_A(C | f, m)$ equals that of algorithm $B$, $\E_B(C | f, m)$, without violating NFL.
In this case the maximal difference between the expected $f$-conditioned OTS errors of the algorithms as one varies over $f$ is zero.
On the other hand, it may instead be that the two algorithms are \textit{anti}-correlated as one varies
over $f$, again, without violating NFL. In other words, it may be that algorithm $A$ performs {better} than random guessing on a function $f$ iff algorithm
$B$ performs {worse} than random guessing on that function. In this case, the maximal
difference between $\E_A(C | f, m)$ and $\E_B(C | f, m)$ as one varies over $f$'s can be quite large. 
As a third possibility, it may be that there are a few $f$ for which algorithm $A$ performs vastly better than
algorithm $B$, but on the large number of other functions $f$, algorithm $B$ performs just slightly better
than algorithm $A$.

%These possibilities mean that despite NFL,
%in general there may be what were called ``head-to-head minimax'' distinctions in~\cite{woma97} that relate the OTS performance
%of a given pair of learning algorithms. 
To make this more formal, fix two learning algorithms $\A_1, \A_2$, producing hypotheses
$h_{\A_1}$ and $h_{\A_2}$, respectively,  and write the 
associated cost functions as $C_{\A_1} = C(f, h_{\A_1}, d)$ and  $C_{\A_2} = C(f, h_{\A_2}, d)$, respectively. NFL tells
us that $\sum_f P(C_{\A_1} \,|\, f, m) = \sum_f P(C_{\A_2} \,|\, f, m) $, i.e., the two marginalizations of
$\sum_f P(C_{\A_1}, C_{\A_2} \,|\, f, m)$ are identical. Nonetheless, 
in general,  $\sum_f P(C_{\A_1}, C_{\A_2} \,|\, f, m)$ need not be a symmetric function
of its two arguments --- it may change if we interchange $\A_1$ and $\A_2$,
i.e., if we redefine $\A_1$ to always produce the hypothesis $h_{\A_2}$ that the original algorithm $\A_2$ would produce
for a given training set $d$, and redefine $\A_2$ to always produce the hypothesis $h_{\A_1}$ that the original algorithm $\A_1$ would
produce on $d$. In this case, we say that there are ``head-to-head minimax'' distinctions~\cite{woma97} between the OTS performances
of the two learning algorithms.
Such distinctions might have had substantial repercussions for co-evolutionary scenarios like the development of life on Earth
under natural selection, as elaborated in~\cite{woma05}.
As described below, it is precisely such head-to-head minimax distinctions that underlie the power of OLEA algorithms \textit{in toto}.

\section{Counter-intuitive implications of the NFL theorems }

In this section I first give a cursory sketch of how the NFL theorems can be consistent with the results
of computational learning theory. I then present
a scenario that illustrates how to exploit the NFL theorems in specific scenarios to derive counter-intuitive
results that do not hold more generally.
\begin{enumerate}

\item Suppose that $P(h \,|\, d) = \delta(h, h^*)$, where $\delta(., .)$ is the Kronecker delta function. So
the learning algorithm always produces the same hypothesis $h^*$, no matter what $d$ is. Also suppose that the likelihood
function is $P(d \,|\, f)$ is noise-free. Consider the data-blind cost function $C(f, h) = \sum_x \pi(x) L(f(x), h(x))$, and define
the associated $f$-conditioned expected empirical cost as
\eq{
\hat{C}(f, m) &:= \sum_{h, d : |d_X| = m, q \in d_X} \dfrac{ P(h \,|\, d) L(f(q), h(q)) P(d \,|\, f) \pi(q)}{ \sum_{q \in d_X}  \pi(q)} \\
	&=  \sum_{d : |d_X| = m, q \in d_X} \dfrac{L(f(q), h^*(q)) P(d_X \,|\, f)\pi(q)}{\sum_{q \in d_X}  \pi(q)}
}
The law of large numbers assures us that $\E(C \,|\, f, m)$ converges to $\hat{C}(f, m)$
as $m$ and $|X|$ both grow with $|X| \gg m$, e.g., for a uniform distribution $\pi$.
Therefore $\E(C \,|\, m)$ as well converges to  $\E(\hat{C} \,|\, m)$, the $m$-conditioned empirical cost. (Note this is true for any prior $P(f)$.) 
One of the primary concerns of the field of computational learning theory is characterizing 
the precise form of this kind of convergence in different scenarios.

Next, note that it's also the case that for
%for any randomly chosen $d_X$ of size $m$, 
for $|X| \gg m$,
$C(f, h^*)$ converges to $\E(C_{OTS} \,|\, f, h^*, m)$, e.g., using the OTS cost function of \cref{eq:OTS_example} for a distribution $\pi$
that is uniform over $X$.
Combining, intuition might lead one to suppose that if $|X| \gg m$, then $\E(C_{OTS} \,|\, m, \hat{C})$ would also become
peaked about $C_{OTS} = \hat{C}(f, m)$. In other words, intuition might suggest that OTS cost converges to the empirical (on-data set)
cost as the size of the space grows and the size of the training set grows. In fact though, by NFL, the average over priors $P(f)$ of
$\E(C_{OTS} \,|\, m, \hat{C})$ is \textit{independent} of $\hat{C}$, since $\hat{C}$ has no statistical coupling
with $C_{OTS}$ under that average. So this intuition is fallacious.

The interested reader is directed to~\cite{adam2019no} for 
further discussion reconciling the NFL theorems and computational learning theory.

\item Given any fixed set of learning algorithms, $\{\A_i\}$, define $\X(\{\A_i\})$ to be the learning algorithm
that for any data set $d$ determines which of the $\A_i$ has lowest cross-validation error on $d$ and then
uses that $\A_i$ to predict the output for all questions $q \not \in d_X$, with any convenient tie-breaking mechanism. 
(For current purposes, there is no need to specify the
precise type of cross-validation, e.g., $K$-fold, leave-one-out, etc.) Similarly define $\XX(\{\A_i\})$ to be the learning algorithm
that for any data set $d$ determines which of the $\A_i$ has \textit{highest} cross-validation error on $d$ and then
uses that $\A_i$ to predict the output for all questions $q \not \in d_X$, with any convenient tie-breaking mechanism.

I will refer to $\X(\{\A_i\})$ and $\XX(\{\A_i\})$ as the ``method of cross-validation'' and the ``method of anti-cross-validation'', respectively.
Note that for any fixed  set of learning algorithms $\{\A_i\}$ that those two methods are applied to, each of them is itself
a learning algorithm, i.e., a map from a provided training set $d$ to a hypothesis function $h$.

Next, suppose that $Y = \B$. Again suppose that the likelihood is noise-free.
 For simplicity consider the case where $\{\A_i\}$ contains exactly two learning algorithms.
The ``majority'' learning algorithm $\A_1$ predicts $1 / 0$ for all off-training set queries, depending
on whether the output $y = 1 / 0$ was more common in the data set (with an arbitrary tie-breaking choice).
The ``anti-majority'' learning algorithm $\A_2$ instead predicts $1 / 0$ for all off-training set queries, depending
on whether the output $y = 0 / 1$ was more common in the data set. (So $\A_1$ predicts whatever was the most common
output in the training set, independent of the precise question $q \not \in d_X$, and $\A_2$ predicts whatever was the 
least common output.) Choose the OTS zero-one cost function,
for simplicity defined for a uniform sampling distribution over the OTS $q$'s: 
\eq{
C_{OTS}(f, h, d) &=\sum_{x \not \in d_X} \dfrac{1 - \delta(h(x), f(x))} {|X| - m}
} 

Consider the prior $P^\dagger(f)$ that allows just the two constant functions: $f(x)=1 \; \forall x \in X$, and 
$f(x)=0 \; \forall x \in X$, assigning each of them the probability $1/2$ . For either of those two constant functions, for any training set $d$,
%whose size is odd (to avoid the possibility of ties), 
$\XX(\{\A_1, \A_2\})$ always makes the wrong prediction for any OTS question.
So the expected OTS zero-one loss of anti-cross-validation is $1$. This is true
whether we condition on a single data set  (as in Bayesian statistics)
or average over all data sets of a given size (as in sampling theory statistics).\footnote{It is also true
%whether we  average over the two allowed $f$'s (as in Bayesian statistics and the NFL theorems), or instead
if we condition on a particular one of the two allowed $f$'s, as in sampling theory statistics, in which
case the prior is irrelevant, and NFL does not apply.} By NFL, this means that there must be some \textit{other}
prior, $P^*(f) \ne P^\dagger(f)$, for which the expected OTS zero-one cost of anti-cross-validation is less than $1/2$.
(Note that in general such a ``compensating'' prior may assign nonzero probability to functions $f$ that are not
constant over all $X$.)

Next, note that the sum of the expected OTS zero-one cost of
cross-validation and anti-cross-validation conditioned on $d$ and one of the two allowed target functions $f$ is independent of $f, d$:
\eq{
\E_{\X(\{A_i\})} \left(C_{OTS} \,|\, d, f \right) + \E_{\hat \X(\{A_i\})} \left(C_{OTS} \,|\, d, f \right) &= 1
}
(This is due to the nature of the majority and anti-majority algorithms and has nothing to do with NFL.)
%which would allow the expected performance of $\X$ and $\XX$ to be positively
%correlated.)
Therefore 
%the sum of the expected errors conditioned on $f, m$ (as in sampling theory statistics) equals $1/2$.
%Similarly, 
for \textit{any} prior $P(f)$ over the two constant functions, the sum of the expected errors conditioned on $d$ (as in Bayesian statistics) equals $1$:
\eq{
\E_{\X(\{A_i\})} \left(C_{OTS} \,|\, d \right) + \E_{\hat \X(\{A_i\})} \left(C_{OTS} \,|\, d \right) &= 1
}
%\eq{
%\E\left(C \,|\, \X(\{A_i\}, m, f\right) + \E\left(C \,|\, \XX(\{A_i\}, m, f \right) 
%	&= \E\left(C \,|\, \X(\{A_i\}, d\right) + \E\left(C \,|\, \XX(\{A_i\}, d\right) = 1/2
%}

Combining, since the expected OTS zero-one loss of anti-cross-validation is less than $1/2$ 
for the prior $P^*(f)$, the expected OTS zero-one loss of cross-validation must
be greater than $1/2$ for that prior. Moreover,
no matter what $f$ is, and therefore no matter the prior, expected zero-one OTS loss is $1/2$ for the algorithm that always
guesses randomly, with probability $1/2$ of choosing $y = 1$. 
%We already knew that since there were $f$'s for which $\XX$ performs better than that random guesser,
%there must be $f$'s for which it performs worse than random. Similarly, we knew that there were $f$'s
%for which $\XX$ performs better than a random coin toss. 

The NFL theorems already told us that there must be a prior $P'(f)$ for which cross-validation performs worse than
random guessing, and that there must be a prior $P''(f)$ for which anti-cross-validation performs \textit{better}
than random guessing. However, one might have suspected that in general, those would have to be different priors, i.e., 
that $P'(f) \ne P''(f)$. In other words, one might have supposed that any prior $P'(f)$ for which anti-cross-validation
does worse than random guessing is also a prior for which cross-validation does worse than random
guessing (and vice-versa).

The analysis above shows that this is not the case:
For the single prior $P^*(f)$, 
%there is a method that is successful, in the sense of performing
%better than random guessing. However, that successful method is 
the method of anti-cross-validation  is successful, in the sense of performing
better than random guessing. However, for that same prior 
%; for that prior 
the method of cross-validation
is \textit{not} successful, and performs worse than a random coin-toss. 
(Note that the experiments recounted in Sec.\,9 of~\cite{schurz2019hume} are consistent with this phenomenon.)
%The head-to-head distinctions between learning algorithms discussed in the previous section in abstract terms
%are concerned with precisely this kind of comparison of learning algorithms on the same prior.

%(Q: Do we need to formulate these kinds of results
%in terms of a prior $P^*(f)$, or can we establish results like these conditioned on some target function $f^*$?))
%
%Then NFL tells us that ...

This kind of phenomenon holds more generally. For \textit{any} method $\I$, and associated ``anti-'' method $\II$, 
there exist priors for which $\II$ is successful, but $\I$ is not. (Or as others might put it, for any such ``meta-induction algorithm'' 
$\I$ and ``anti-meta-induction algorithm'' $\II$.)
In this very specific sense, any claim that some such method $\I$  ``is guaranteed to be successful, no matter the
course of nature, if any method is''~\cite{sterkenburg2019meta} is wrong. This is true even though there \textit{can} be 
a method $\I$ that performs better than the associated method $\II$ in head-to-head minimax distinctions, 
``no matter the course of nature''. In sum, whether one method can have such guarantees 
over another depends on how precisely one is comparing 
methods.
\end{enumerate}

\section{No free lunch and OLEA}

The central problem in the simplified OLEA scenario introduced in \cref{sec:introduction} is how to set the value $v_{K+1}(m)$ based
on knowledge of the values of the preceding values of the $K$ other sequences, $\{v_k(i) : i \in \{1, \ldots, m-1\}, k \in \{1, \ldots, K\}\}$.
One can map this problem into a special case of the 
supervised learning problem of how to generalize from a particular training set $d$. Take $X = \Z^+$, and identify the successive
iterations $i \in \Z^+$ with successive elements of $X$. Also take $Y = \B$.
So each $f$ is a function from $\Z^+ \rightarrow \B$. 
Identify $d_X$ with the first $m$ counting numbers, and choose $d_Y$ to be any vector of $m$ bits. Also identify
each of the first $K$ sequences $\{v_k(x) : k = 1, \ldots, K\}$ 
with the values of $K$ different \textbf{considered} functions, $\{g_k(x) : k = 1, \ldots, K\}$,
by setting $g_k(x) = d_Y(x)$ iff $v_k(x) = 1$, for all $x \in d_X$. 
(The values of those candidate functions for values $x > m$ is arbitrary.)
%So a given candidate function $g_k$ equals $0/1$ for a given $x$ depending on whether 
So the sequence $v_k$ has a payoff value of $0 / 1$ for iteration $x$ depending on whether $g_k$ agrees with the
training set on $x \in d_X$. Assume a noise-free likelihood
function $P(d \,|\, f)$, and adopt a OTS cost function $C(f, h, d) = 1 - \delta(h(|d_X| + 1), f(|d_X| + 1))$.

At heart, when the algorithms considered in~\cite{schurz2019hume} are mapped this way into the realm of supervised learning,
they become various learning algorithms for using the $m$-element training set $d$ to combine the values of the
candidate functions $g_k$ evaluated for $x = m +1$ in order to set the value of the hypothesis function at $x = m+1$,
i.e., in order to set $h(m + 1)$. The associated value $v_{K+1}(m+1)$ is
set to $0 / 1$ depending on whether that value $h(m+1)$ produced from the candidate functions
equals $f(m+1)$. So translated back into the context of~\cite{schurz2019hume}, the OTS cost function is $0 / 1$
depending on whether $v_{K+1}(x) = 0 / 1$.
%~\footnote{Strictly speaking, the 
%algorithms considered in~\cite{schurz2019hume} don't provide full hypothesis functions $h(x)$ defined
%over all $X$, but only provide the value $h(x)$ for the $x$ immediately following the last element in
%the training set. This simply corresponds to a slightly more complicated choice of OTS cost function though, and doesn't affect
%the arguments in this section.}

The NFL theorems for supervised learning tell us immediately that averaged over all $f$, the expected value of this OTS cost
is $1/2$ for \textit{all} algorithms $v_{K+1}$ (not just those algorithms considered in~\cite{schurz2019hume}). 
More generally, if we average uniformly over all priors $P(f)$, then the expected 
value of $v_{K+1}(x)$ in any iteration $x$ is $1/2$. This is true \textit{no matter how 
we set the sequence $v_{K+1}$}, no matter what $d$ is, and no matter what the candidate functions ${g_k}$ are,
i.e., no matter what the sequences $\{v_k : k = 1, \dots, K\}$ are.

How can these NFL results for OLEA be reconciled with the regret-reducing results of OLEA in general, and
the benefits of AW algorithms in particular? The answer was 
provided in~\cite{woma97}: OLEA results concern head-to-head 
distinctions between learning algorithms, and there can be free lunches for head-to-head distinctions.
Whether such free lunches are normative, determining how one ``should'' make predictions is a nuanced
topic. (For example, recall the discussion above about natural selection and co-evolutionary free lunches.)
Under the most conventional formulations of Bayesian decision theory, the answer is 'no', these
kinds of distinctions do not provide a reason to prefer one algorithm over another. In this sense, Schurz's
claim to ``solve Hume's problem'' results from subtle and rich but ultimately flawed reasoning. 

As a final point, while Schurz does not consider the NFL theorems involving an average over priors $P(f)$, he does address
the case of a uniform $P(f)$, by contesting its ``epistemic validity''. Specifically Schurz argues that one ``should'' adopt a 
single, specific prior, in a normative sense (a stance I do not promote). However, Schurz argues that
it should be a uniform prior over \textit{frequencies} of the future sequences of bits, 
rather than (as under uniform $P(f)$) over the
patterns of those bits. 

In response to this it is important to point out that \textit{all} of statistical physics is based on 
a uniform distribution over patterns, \textit{not} over frequencies; that uniform distribution over patterns
is known as the \textbf{microcanonical ensemble}. As an example, under the microcanonical ensemble, the
distribution of joint states of all the binary spins in an Ising spin is uniform, and so the distribution of frequencies
of the average spin value is highly \textit{non}-uniform. Indeed, the whole validity of standard, macroscopic thermodynamics,
relies on the fact that in the thermodynamic limit of an infinite number of spins, the distribution
over frequencies becomes a Dirac delta function.

In addition, the highly successful Maximum entropy procedure 
for inductively inferring (!) a probability distribution from knowledge of its moments relies on a uniform
prior over patterns, not over frequencies~\cite{jabr03,jaynes1968prior}. In short,
Schurz's proposal for a uniform prior over frequencies runs afoul of thousands (tens of thousands?) of previous experiments
concerning the real, physical world. Again, the central issue is how one is comparing algorithms. In all
of those real-world experiments, the key issue is not head-to-head minimax distinctions, 
which is what allows there to be such strong arguments in favor of a uniform prior.

$ $

\noindent \textbf{Acknowledgements.} I would like to thank the Santa Fe Institute for support.

\noindent \bibliographystyle{unsrt}
%\bibliography{../../../../BIB/refs}

\end{document}